\documentclass{article}
\usepackage{spconf,amsmath,amssymb,graphicx,hyperref}
\usepackage{xurl}
\hypersetup{hidelinks}
\usepackage[utf8]{inputenc}
\usepackage[T5,T1]{fontenc}
\usepackage{textcomp}
\usepackage{CJKutf8}
\newcommand{\zh}[1]{\begin{CJK}{UTF8}{gbsn}#1\end{CJK}}
\newcommand{\mubench}{mu-bench}
\usepackage{booktabs}
\usepackage{float}

\usepackage{tabularx}
\usepackage{array}
\usepackage{enumitem}
\setlist{leftmargin=1.5em,itemsep=1pt,topsep=2pt,parsep=0pt}

\title{MU-BENCH: A \underline{M}ULTILINGUAL \underline{U}TTERANCE TRANSCRIPTION BENCHMARK}
\twoauthors
  {Andrea Li\thanks{Authors listed in alphabetical order.}}
    {UC Berkeley}
  {Soham Ray}
    {Sierra AI}
\begin{document}
\ninept
\maketitle
\begin{abstract}
Voice agents depend on accurate automatic speech recognition (ASR) to act on what callers say, yet ASR is evaluated on read, English-centric speech with word error rate (WER), which penalizes surface rather than semantic differences. We introduce mu-bench, a dataset of 4,270 caller utterances from 250 phone calls to an AI banking agent in English, Spanish, Turkish, Vietnamese, and Mandarin, centered on form-field inputs such as names, email addresses, and confirmation codes. We release Utterance Error Rate (UER), an LLM judge of whether a transcript preserves meaning, calibrated against human raters, together with an LLM normalizer that makes WER comparable across providers' output formats. On 1,847 human-rated transcripts, UER agrees with annotators at $\kappa$ = 0.78, versus 0.53 for exact-match WER on normalized text. We rank six commercial providers on a public leaderboard; the best reaches 11.9\% UER, and Mandarin is hardest for all six.
\end{abstract}
\section{Introduction}
\label{sec:intro}

AI agents are increasingly deployed for voice interactions such as booking appointments, processing returns, disputing charges, and refilling prescriptions. Accurate automatic speech recognition (ASR) across languages is therefore critical: transcription output now shapes the actions taken on a customer's behalf.

We introduce \mubench, a multilingual utterance transcription benchmark of phone calls between humans and an AI banking agent in English, Spanish, Turkish, Vietnamese, and Mandarin. We make three contributions:
\begin{enumerate}
  \item \textbf{Dataset}: Multilingual, agent-to-human phone conversations centered on form-field input collection, released on Hugging Face as \texttt{sierra-research/}\allowbreak\texttt{mu-bench}.
  \item \textbf{Evaluation metric}: Utterance Error Rate (UER), an LLM judge calibrated against human judgment in five languages, and an LLM normalizer that aligns text.
  \item \textbf{Leaderboard}: Publically available at \url{https://research.sierra.ai/mubench/} where providers can submit transcripts and compare models.
\end{enumerate}
All code and prompts are at \url{https://github.com/sierra-research/mu-bench/tree/icassp-2027}.

\section{Related Work}
\label{sec:related}

\subsection{Benchmarks and leaderboards}
\label{ssec:gaps}

There are three properties important to voice agents that no existing benchmark covers all of.
\begin{itemize}[nosep, leftmargin=*]
  \item \textbf{Multilingual:} only about 18\% of the world's population speaks English \cite{Eberhard2026, UnitedNations2024}, yet the corpora used to train and evaluate frontier ASR models remain overwhelmingly English-only.
  \item \textbf{Conversational:} live phone audio contains disfluencies, interruptions, and channel noise, and differs materially from volunteers reading scripts into a microphone.
  \item \textbf{Form-field input:} agent calls routinely collect names, account numbers, addresses, and confirmation codes, where a single misheard token is consequential.
\end{itemize}
Read-speech corpora such as LibriSpeech \cite{Panayotov2015}, MLS \cite{Pratap2020}, Common Voice \cite{Ardila2020}, and FLEURS \cite{Conneau2023} are multilingual but not conversational; conversational corpora such as Switchboard \cite{Godfrey1992} and CORAAL \cite{Kendall2018} are English-only. Multilingual conversational datasets, CallHome \cite{Canavan1996} and Fisher Spanish \cite{Graff2010} don't target input collection (Table~\ref{tab:benchmarks}). 

Additionally, leaderboards such as the Hugging Face Open ASR Leaderboard \cite{Srivastav2023} and Artificial Analysis \cite{ArtificialAnalysis2026} pool several of these datasets; \mubench{} aims to complements these results rather than replace them.

\subsection{Metrics and text normalization}
\label{ssec:correctness}

The standard ASR metric is word error rate (WER): the word-level Levenshtein distance defined as $\text{WER} = \frac{S + D + I}{N}$,
where $S$, $D$, and $I$ are substitutions, deletions, and insertions and $N$ is the reference length. Every edit weighs the same regardless of meaning: against ``this is a cat,'' both ``this is the cat'' and ``this is a cap'' score 25\%, yet only the latter changes the meaning.

This has motivated ``semantic WER'' variants, from embedding-based SemDist \cite{Kim2021} to LLM-based rubrics such as LASER \cite{Parulekar2025}. LASER prompts an LLM with a three-tier penalty rubric (none, minor, major) and worked Hindi examples. Table~\ref{tab:kappa} reports LASER's prompt run on \mubench{}, which penalizes filler-word differences that \mubench{} does not.. 

A second challenge is normalizing hypotheses and references before scoring. The de facto standard, the Whisper normalizer \cite{Radford2022}, applies English-specific rules (removing fillers, standardizing contractions) but only basic handling elsewhere. \mubench{} releases an open LLM-based normalizer covering multilingual and form-field text (\S\ref{ssec:normalization-method}).

\begin{table}[t]
\centering
\small
\caption{Coverage of ASR benchmarks on the three criteria in \S\ref{ssec:gaps}}
\label{tab:benchmarks}
\vspace{4pt}
\setlength{\tabcolsep}{3pt}
\begin{tabular}{lcccl}
\hline
\textbf{Benchmark} & \textbf{Multi.} & \textbf{Conv.} & \textbf{Form} & \textbf{Source} \\
\hline
LibriSpeech \cite{Panayotov2015}      &            &            &            & Read \\
TED-LIUM 3 \cite{Hernandez2018}       &            &            &            & Prepared \\
GigaSpeech \cite{Chen2021}            &            &            &            & Web audio \\
SPGISpeech \cite{ONeill2021}          &            &            &            & Earnings calls \\
People's Speech \cite{Galvez2021}     &            &            &            & Found \\
MLS \cite{Pratap2020}                 & \checkmark &            &            & Read \\
Common Voice \cite{Ardila2020}        & \checkmark &            &            & Read \\
FLEURS \cite{Conneau2023}             & \checkmark &            &            & Read \\
CoVoST 2 \cite{Wang2021}              & \checkmark &            &            & Read \\
CORAAL \cite{Kendall2018}             &            & \checkmark &            & Interviews \\
Switchboard \cite{Godfrey1992}        &            & \checkmark &            & Phone calls \\
CallHome \cite{Canavan1996}           & \checkmark & \checkmark &            & Phone calls \\
Fisher Spanish \cite{Graff2010}       &            & \checkmark &            & Phone calls \\
\hline
\mubench                              & \checkmark & \checkmark & \checkmark & Phone calls \\
\hline
\end{tabular}
\end{table}

\section{Methods and Approach}
\label{sec:methods}

\subsection{Dataset}
\label{ssec:dataset}

\noindent\textbf{Callers.}\label{sssec:collection} Native speakers called an AI banking agent, role-playing a fictional customer with one of four personas: frustrated, polite, casual and indifferent, or overly detailed. Calls used callers' own phones and environments, giving natural background noise, at 8~kHz mono over the telephony channel.

\noindent\textbf{AI banking agent.} The agent follows the standard voice-agent pipeline: speech-to-text, an LLM that reasons over the transcript and replies, and text-to-speech. It authenticates the caller by collecting a name, email, or phone number, checks card status, raises credit limits, disputes transactions, and issues a case tracking code via a tool call. Only the caller side is released; agent turns are synthetic speech and are discarded.

\noindent\textbf{Ground truth.}\label{sssec:annotation} Professional native-speaking annotators produced ground truth in two passes: one segmented each call by speaker turn and wrote a clean verbatim transcript with timestamps; a second independently verified and corrected it. Recordings are then split into caller turns at these timestamps, and turns flagged unintelligible are excluded.\label{sssec:processing} Table~\ref{tab:stats} summarizes the release.

\begin{table}[t]
\centering
\caption{\mubench{} composition by locale. Length is the mean utterance length in seconds.}
\label{tab:stats}
\vspace{4pt}
\setlength{\tabcolsep}{4pt}
\begin{tabular}{lrrrr}
\hline
\textbf{Locale} & \textbf{Utt.} & \textbf{Conv.} & \textbf{Audio (hr)} & \textbf{Length (s)} \\
\hline
en-US & 817 & 50 & 0.93 & 4.1 \\
es-MX & 792 & 50 & 1.01 & 4.6 \\
tr-TR & 846 & 50 & 1.18 & 5.0 \\
vi-VN & 975 & 50 & 1.21 & 4.5 \\
zh-CN & 840 & 50 & 0.76 & 3.3 \\
\hline
Total & 4{,}270 & 250 & 5.08 & 4.3 \\
\hline
\end{tabular}
\end{table}

\subsection{Language-aware normalization}
\label{ssec:normalization-method}

Given a reference and a predicted transcript, the LLM normalizer reformats the prediction into the rules of the reference so that stylistic variance across providers is scored fairly. We call this \emph{reference-guided} normalization as opposed to blind normalization, because the LLM is given the ground truth as reference. The reference is needed because spoken-form mappings are one-to-many: ``52'' may have been said as \emph{five two} or \emph{fifty-two}, and only the reference says which without having to enumerate rules. The price is that a rewrite toward the reference can hide a real error; \S\ref{ssec:coverage} measures that leak and compares against reference-blind normalization. Table~\ref{tab:whisper-gaps} lists the rule-based failure modes the normalizer covers. Non-English structure adds more, such as Mandarin homophones that audio alone cannot disambiguate: \zh{我叫羽凡} and \zh{我叫宇凡} are both pronounced \emph{w\v{o} ji\`ao y\v{u} f\'an} (``My name is Y\v{u}f\'an''), so choosing \zh{羽} over \zh{宇} should not be penalized. Yet homophones cannot be normalized blindly: \zh{买卖} ('business') and \zh{买麦} ('buy wheat') share the pronunciation \emph{m\v{a}i m\`ai}. The normalizer's effect on agreement with humans is analyzed in \S\ref{ssec:rq2} and its coverage in \S\ref{ssec:coverage}.

\subsection{Utterance Error Rate}
\label{ssec:uer}

Even with perfect normalization, WER penalizes semantically equivalent cases such as ``hi'' versus ``hey.'' We therefore give the normalized prediction and the reference to an LLM judge with the rubric:
\begin{itemize}
  \item \textbf{Score 1, significant.} The meaning is derailed or incoherent relative to the reference, or any component of a form-field input is misspelled.
  \item \textbf{Score 2, minor.} Some words differ, but the meaning of the overall sentence does not change.
  \item \textbf{Score 3, no error.} The utterances are semantically the same.
\end{itemize}

UER is the fraction of utterances rated as containing a significant error, averaged without weighting over the $L=5$ locales:
\begin{equation}
\text{UER} = \frac{1}{L} \sum_{\ell=1}^{L} \frac{|\{u \in \mathcal{U}_\ell : s(u) = 1\}|}{|\mathcal{U}_\ell|},
\label{eq:uer}
\end{equation}
where $\mathcal{U}_\ell$ is the set of utterances in locale $\ell$ and $s(u)=1$ if the judge rated $u$ as a significant error.

\section{Experimental Setup}
\label{sec:setup}

\begin{table}[!t]
\centering
\small
\caption{Cohen's $\kappa$ against human annotators on the binary label \emph{significant error} vs.\ not, sorted by $\kappa$. Norm = normalization; blind LLM is the same \texttt{gpt-6-astra} normalizer given one transcript at a time (\S\ref{ssec:coverage}). WER$_{>0}$ flags any difference after normalization. LASER is the published rubric of \cite{Parulekar2025} run verbatim with the same \texttt{gpt-6-astra} judge, flagging any major-penalty error.}
\label{tab:kappa}
\vspace{4pt}
\setlength{\tabcolsep}{3pt}
\begin{tabular}{llcccc}
\hline
\textbf{Norm.} & \textbf{Metric} & $\boldsymbol{\kappa}\,\uparrow$ & \textbf{95\% CI} & \textbf{Prec}\,$\uparrow$ & \textbf{Rec}\,$\uparrow$ \\
\hline
\mubench & UER & \textbf{0.779} & [0.717, 0.835] & 0.859 & 0.764 \\
none        & UER & 0.764 & [0.699, 0.821] & 0.831 & 0.764 \\
Whisper     & UER & 0.681 & [0.606, 0.748] & 0.705 & 0.753 \\
blind LLM   & UER & 0.646 & [0.567, 0.716] & 0.690 & 0.703 \\
\mubench & LASER & 0.556 & [0.475, 0.634] & 0.514 & 0.825 \\
\mubench & WER$_{>0}$ & 0.533 & [0.453, 0.613] & 0.462 & 0.954 \\
none        & LASER & 0.460 & [0.383, 0.535] & 0.424 & 0.833 \\
Whisper     & WER$_{>0}$ & 0.260 & [0.209, 0.315] & 0.271 & 1.000 \\
none        & WER$_{>0}$ & 0.098 & [0.076, 0.124] & 0.187 & 1.000 \\
\hline
\end{tabular}
\end{table}

\begin{table*}[!t]
\centering
\small
\caption{Effect of model size on the normalizer and the judge, varied one at a time; mean over five seeds (sd across seeds at most 0.011 for Overall and 0.048 for any locale cell). Flip is the share of rows whose UER verdict differs between seeds of the varied component (the other is fixed at seed 7, hence two different \texttt{gpt-6-astra} rows). Prec = precision, Rec = recall, \$/1k = list price per 1{,}000 rows.}
\label{tab:size-sweep}
\vspace{4pt}
\setlength{\tabcolsep}{3pt}
\begin{tabular}{lrcccccccc c}
\hline
 & & \multicolumn{6}{c}{\textbf{UER} $\boldsymbol{\kappa}\,\uparrow$} & & & \\
\cmidrule(lr){3-8}
\textbf{Model} & \textbf{\$/1k} & \textbf{Overall} & en-US & es-MX & tr-TR & vi-VN & zh-CN & \textbf{Prec} & \textbf{Rec} & \textbf{Flip (\%)}\,$\downarrow$ \\
\hline
\multicolumn{11}{l}{\textit{Normalizer varied}} \\
\texttt{gpt-5.6-luna}  & 0.22 & 0.755  & \textbf{0.745}  & \textbf{0.687} & 0.698 & \textbf{0.840} & 0.758 & 0.881 & 0.710 & 1.0 \\
\texttt{gpt-5.6-terra} & 1.62 & 0.704  & 0.699  & 0.647 & 0.598 & 0.830  & 0.701 & 0.835 & 0.666 & 1.4 \\
\texttt{gpt-5.6-sol}   & 3.08 & 0.751  & 0.743 & 0.667  & 0.709 & \textbf{0.840} & 0.754 & 0.851 & 0.726 & 0.5 \\
\texttt{gpt-6-astra}   & 7.39 & \textbf{0.775} & 0.721 & 0.654 & \textbf{0.717} & \textbf{0.840} & \textbf{0.839} & 0.861 & 0.757 & \textbf{0.0} \\
\hline
\multicolumn{11}{l}{\textit{Judge varied}} \\
\texttt{gpt-5.6-luna}  & 0.04 & 0.682 & 0.603 & 0.675 & 0.600 & 0.742 & 0.700 & 0.929 & 0.584 & 1.8 \\
\texttt{gpt-5.6-terra} & 0.33 & 0.682 & 0.653 & 0.595 & 0.645 & 0.750 & 0.691 & 0.894 & 0.601 & 1.4 \\
\texttt{gpt-5.6-sol}   & 0.77 & \textbf{0.774} & 0.666 & \textbf{0.678} & 0.693 & \textbf{0.866} & 0.844 & 0.840 & 0.773 & 1.0 \\
\texttt{gpt-6-astra}   & 1.62 & \textbf{0.774} & \textbf{0.684} & 0.657 & \textbf{0.719} & 0.829 & \textbf{0.852} & 0.852 & 0.762 & \textbf{0.3} \\
\hline
\end{tabular}
\end{table*}

\begin{table*}[!t]
\centering
\caption{Results on \mubench, sorted by UER (\%). IC = UER on the 1{,}706 input-collection utterances (\S\ref{ssec:input-collection}), macro-averaged over locales like \textbf{all}.}
\label{tab:leaderboard}
\vspace{4pt}
\setlength{\tabcolsep}{5pt}
\begin{tabular}{llcccccccrr}
\hline
 & & \multicolumn{7}{c}{\textbf{UER (\%)} $\downarrow$} & \multicolumn{2}{c}{\textbf{Latency (ms)} $\downarrow$} \\
\textbf{Provider} & \textbf{Mode} & en-US & es-MX & tr-TR & vi-VN & zh-CN & \textbf{all} & IC & p50 & p95 \\
\hline
Google Chirp-3 & batch & 3.2 & 7.3 & 9.5 & 6.7 & 32.9 & \textbf{11.9} & 17.7 & 734 & 1136 \\
Google Chirp-3 & stream & 5.8 & 11.6 & 14.2 & 12.0 & 38.0 & \textbf{16.3} & 24.0 & 646 & 887 \\
Microsoft Azure Speech & batch & 2.8 & 11.2 & 17.5 & 22.1 & 31.2 & \textbf{17.0} & 20.1 & 315 & 1030 \\
ElevenLabs Scribe v2 & batch & 4.8 & 10.5 & 12.3 & 24.2 & 34.8 & \textbf{17.3} & 16.6 & 415 & 847 \\
Microsoft Azure Speech & stream & 3.1 & 15.7 & 16.3 & 18.8 & 41.3 & \textbf{19.0} & 28.9 & 242 & 431 \\
OpenAI GPT-4o Mini Transcribe & stream & 2.9 & 14.5 & 22.0 & 33.5 & 47.4 & \textbf{24.1} & 33.2 & 610 & 1035 \\
xAI Grok STT & batch & 5.8 & 20.2 & 27.3 & 34.3 & 42.5 & \textbf{26.0} & 27.2 & 316 & 1965 \\
OpenAI GPT-4o Mini Transcribe & batch & 4.0 & 17.7 & 22.8 & 40.1 & 51.2 & \textbf{27.2} & 34.7 & 661 & 1117 \\
Deepgram Nova-3 & batch & 3.9 & 11.9 & 20.7 & 50.8 & 52.7 & \textbf{28.0} & 34.0 & 107 & 376 \\
\hline
\end{tabular}
\end{table*}

\begin{table*}[!t]
\centering
\small
\caption{Failure classes of rule-based normalization on semantically equivalent outputs: the pair as transcribed (Raw) and after Whisper normalization, which still leaves it distinct; \mubench{} maps the prediction onto the raw reference. Count is the number of (utterance, provider) pairs only \mubench{} reconciles.}
\label{tab:whisper-gaps}
\vspace{4pt}
\setlength{\tabcolsep}{4pt}
\renewcommand{\arraystretch}{1.15}
\begin{tabularx}{\textwidth}{>{\raggedright\arraybackslash}p{2.3cm} r
  >{\raggedright\arraybackslash}X >{\raggedright\arraybackslash}X
  >{\raggedright\arraybackslash}X >{\raggedright\arraybackslash}X}
\hline
 & & \multicolumn{2}{c}{\textbf{Raw}} & \multicolumn{2}{c}{\textbf{Whisper-normalized}} \\
\cmidrule(lr){3-4}\cmidrule(lr){5-6}
\textbf{Class} & \textbf{Count} & Reference & Prediction & Reference & Prediction \\
\hline
Digits vs.\ words   & 1{,}496 & C N 6 5 3\zh{。} & CN\zh{六五三} & c n 6 5 3 & cn\zh{六五三} \\
Contact format      & 1{,}141 & Ashley dot Brown at email dot com. & ashley.brown@email.com & ashley dot brown at email dot com & ashley brown email com \\
Grouping  & 1{,}058 & 55 12 34 56 7 8. & 55 12 34 56 78 & 55 12 34 56 7 8 & 55 12 34 56 78 \\
Fillers (non-en) &   841   & S\'i. & Este, s\'i. & s\'i & este s\'i \\
Word variant        &   543   & Okay. & OK. & okay & ok \\
Diacritics          &   176   & S\'i. & Si. & s\'i & si \\
Script (zh-CN)      &   108   & \zh{美玲。} & Meiling. & \zh{美玲} & meiling \\
\hline
\end{tabularx}
\end{table*}

\subsection{Evaluated providers and settings}
\label{ssec:provider-runs}

We tested six commercial APIs through their batch endpoints and three through their streaming endpoints (Table~\ref{tab:leaderboard}). Batch sends each utterance as one 8~kHz mono PCM16 WAV file, one request at a time; streaming sends 20~ms PCM16 frames (160 samples at 8~kHz). The raw transcript is captured as-is, the locale is passed as a language hint, and all other parameters are left at their defaults.

\subsection{Latency measurement}
\label{ssec:latency-setup}

We measure wall-clock time from the end of speech to the transcript, from a single client at concurrency 1, and report p50 and p95. For batch this is the request-to-response round trip, including the upload; for streaming it is the time from the last frame to final transcript. Time-to-first-token is not used, so the two modes stay comparable.

\subsection{Model}
\label{ssec:sweep-setup}

Normalizer and judge both use \texttt{gpt-6-astra} with \texttt{seed=7}, at \texttt{reasoning\_effort=\allowbreak high} and \texttt{low} respectively; \S\ref{ssec:model-sweep} evaluates alternatives.

\subsection{Human annotation study}
\label{ssec:annotation-study}

We randomly sampled 300 utterances, 60 per locale, in three equal strata of no error, minor error, and significant error. This resulted in 1{,}847 rated rows. One native speaker per locale rated each row against the reference as \emph{no error}, \emph{minor error}, or \emph{significant error}, with provider names hidden and order randomized, using the same category descriptions as the UER judge.

\section{Results}
\label{sec:results}

\subsection{Which normalization and evaluation metric agree best with human judgment?}
\label{ssec:rq2}

Table~\ref{tab:kappa} shows that UER on LLM-normalized text agrees best with human judgment and degrades far less than WER when the text is raw or Whisper-normalized; reference-blind LLM normalization (\S\ref{ssec:coverage}) falls below raw text. WER$_{>0}$ has recall near 1 but precision of 0.19--0.46: it flags surface differences as errors. Even a WER threshold tuned on these same rows, an optimistic bound, reaches only $\kappa=0.67$ on \mubench-normalized text (0.65 when the threshold is chosen on four locales and scored on the fifth), 0.40 on Whisper-normalized and 0.33 on raw text. The normalizer's gain for UER itself is within noise (0.779 vs.\ 0.764; paired difference 0.015, 95\% CI $[-0.010, 0.044]$) and it raises precision from 0.83 to 0.86; we keep it because it makes WER usable (0.53 vs.\ 0.10) and powers the diagnostics of Table~\ref{tab:whisper-gaps}.

\subsection{How much does the underlying LLM model matter?}
\label{ssec:model-sweep}

Table~\ref{tab:size-sweep} varies the normalizer and judge model one at a time over five seeded runs. Size matters far more for the judge: swapping the normalizer from \texttt{gpt-6-astra} to the smallest model, \texttt{gpt-5.6-luna}, costs 0.02 $\kappa$ (0.775 to 0.755) at 1/34 of the price, whereas the same swap on the judge costs 0.09 (0.774 to 0.682). The gap is recall: the smaller judges match or exceed \texttt{gpt-6-astra}'s precision (0.93 and 0.89 vs.\ 0.85) but their recall drops from 0.76 to 0.58 and 0.60, i.e.\ they miss errors rather than invent them. These gaps are not sampling noise: across seeds the pooled $\kappa$ of every cell has a standard deviation of at most 0.011. For the reported configuration, 0.3\% of verdicts change between seeds, and although the normalizer alters its output on 1.1\% of rows, none of those changes flips a verdict.

\subsection{How do frontier providers perform?}
\label{ssec:q1}

Table~\ref{tab:leaderboard} reports the results. English is the strongest locale and Mandarin lags far behind all others. Accuracy and latency are not aligned: Deepgram has the lowest latency but the highest UER, Google the highest median latency and the lowest UER. Streaming and batch endpoints produce different transcripts for many utterances (35\% for Google, 47\% for OpenAI, 70\% for Azure), and switching between them moves macro UER by 2--4 points in either direction (Google $+4.4$, OpenAI $-3.1$).

\subsubsection{Statistical validity}
\label{ssec:significance}

We resample the 250 conversations with replacement 10{,}000 times (utterances within a call share a speaker and channel), recompute every row's UER, and count how often each pairwise ordering flips. Of the 36 pairs among the nine rows, 32 are separated at $p<0.05$ and 27 at $p<0.001$, the latter surviving a Bonferroni correction ($p<0.0014$); the exceptions are Azure batch vs.\ ElevenLabs ($p=0.28$), Google stream vs.\ Azure batch ($0.11$), OpenAI batch vs.\ Deepgram ($0.10$), and Grok vs.\ OpenAI batch ($0.051$). Batch and stream are separated for all three providers run both ways ($p\le0.006$).

We also sent every utterance twice, unchanged and back to back, to measure the APIs' own variance. Google, Azure and Deepgram are deterministic with 0--0.3\% of transcripts changed and no UER movement; ElevenLabs and OpenAI return a different transcript for 20\% and 45\% of identical requests but UER moves by at most 0.4~pp. The leaderboard is stable under resends.

\subsection{How do form-field inputs affect accuracy?}
\label{ssec:input-collection}

We define an utterance as having input-collection if the caller is supplying a value the agent is collecting (a name, email, phone number, ID, code, address, date, or amount). 1{,}706 of 4{,}270 utterances (40\%) are input collection, unevenly across locales: 29\% in English, 35--37\% in Vietnamese and Spanish, and 49--50\% in Mandarin and Turkish. Thus Mandarin is the hardest locale because of input collection; on non-collection utterances, Mandarin's overall UER (27\%) is below Vietnamese (31\%).

Over the six batch providers, input-collection utterances have \emph{lower} normalized WER than the rest (15.0\% vs.\ 18.7\%) yet much \emph{higher} UER (28.6\% vs.\ 17.0\%), as expected from the definition of UER. The exact performance is provider-dependent, where Google's lead is on non-collection speech (6.4\% vs.\ 17.7\% on collected values, Table~\ref{tab:leaderboard}); ElevenLabs is flat (16.6\% on both) and is the best provider on input collection; OpenAI and Deepgram are the worst at about 34\%.

\subsection{What failures does LLM normalization catch?}
\label{ssec:coverage}

Of the 25{,}620 (utterance, provider) pairs, 20.5\% of raw predictions already equal the reference. Of the 20{,}357 that differ as written, Whisper normalization reconciles 23.2\% and \mubench{} 49.3\%, a superset (99.4\% of Whisper's pairs). Table~\ref{tab:whisper-gaps} classifies the 5{,}363 pairs only \mubench{} reconciles.
We accept this at the cost of hiding a small number of real errors: on the human-rated rows it rewrote 12 of the 263 significant errors (4.6\%) into the reference. Judging every pair on raw text as well, no provider's UER moves by more than 0.6~pp and the ranking is unchanged. The LLM alternative, \emph{reference-blind} normalization, canonicalizes each side without seeing the other and reaches only $\kappa = 0.646$ (Table~\ref{tab:kappa}), below raw text. We find that the one-to-many mapping of \S\ref{ssec:normalization-method} now applies to each side independently, so a reference ``52'' becomes \emph{cinco dos} while the prediction's \emph{cincuenta y dos} stays, and the judge sees a content difference. It also raises false alarms on 5.9\% of the human-clean pairs that differ as written and hides 8 significant errors (3\%).

\subsection{Does padding the audio with silence improve quality?}
\label{ssec:padding}

\begin{table}[t]
\centering
\small
\caption{Effects of prepending silence. ``Changed'' is the share of predictions that differ from the unpadded request. Grok was not run: we had no API access at the time.}
\label{tab:padding}
\vspace{4pt}
\setlength{\tabcolsep}{2pt}
\begin{tabular}{l rr @{\hspace{0.3em}} rr @{\hspace{0.3em}} rr}
\hline
 & \multicolumn{2}{c}{\textbf{Changed (\%)}}
 & \multicolumn{2}{c}{$\boldsymbol{\Delta}$\textbf{UER (pp)}}
 & \multicolumn{2}{c}{$\boldsymbol{\Delta}$\textbf{Latency (ms)}} \\
\textbf{Provider} & 1s & 3s & 1s & 3s & 1s & 3s \\
\hline
Deepgram Nova-3   & 20.6 & 21.5 & $-0.1$ & $-0.8$ & $+1$  & $+3$  \\
Azure Speech      & 29.4 & 32.4 & $+1.5$ & $+2.1$ & $+7$  & $+21$ \\
ElevenLabs Scribe & 28.5 & 28.2 & $+0.5$ & $-0.3$ & $+6$  & $+12$ \\
OpenAI 4o Mini    & 36.6 & 37.5 & $-0.6$ & $+0.7$ & $+7$  & $+30$ \\
Google Chirp-3    & 18.4 & 20.1 & $+0.2$ & $+0.2$ & $+18$ & $+57$ \\
\hline
\end{tabular}
\end{table}

Since the average utterance is only 4.3~s (Table~\ref{tab:stats}), we tested prepending 1 or 3~s of digital silence, sending padded and unpadded requests back to back in random order. Padding changes 18--38\% of transcripts but moves UER by at most 2.1~pp for Azure and within 1~pp for the other providers while adding latency in proportion to the audio added (Table~\ref{tab:padding}). \mubench{} therefore clips short audio without padding.

\section{Future Work}
\label{sec:future}

We aim to expand in three directions: more locales (which requires native speakers to review normalization and UER quality against each language's structure), more domain contexts beyond banking, and evaluation of more provider settings such as dynamic keyword boosting, prompting, and noise suppression.

\clearpage
\bibliographystyle{IEEEbib}
\bibliography{refs}

\section{Compliance with Ethical Standards}
This study involved human participants: native speakers who were hired and compensated to place phone calls to an AI banking agent, transcribe audio recordings, or rate transcripts on similarity. All callers gave informed consent to be recorded and to the public
release of their recordings. Callers role-played fictional customers
with invented names, contact details, and account information, so
the released audio and transcripts contain no real personal data.

Data collection was conducted by Sierra AI, which does not maintain a
formal ethics review committee; the study was conducted in accordance
with the 1964 Helsinki Declaration and its later amendments.

\section{Acknowledgments}
This work was supported by Sierra AI. We thank the voice team, in particular Venumadhav Satuluri and Mindy Long, for early work on transcription benchmarking that shaped this paper.
\end{document}